\documentclass[runningheads]{llncs}
\usepackage[T1]{fontenc}
\usepackage{graphicx}%
\usepackage{multirow}%
\usepackage{amsmath,amsfonts,amssymb}
\usepackage{mathrsfs}%
\usepackage[title]{appendix}%
\usepackage{xcolor}
\usepackage{textcomp}%
\usepackage{manyfoot}%
\usepackage{booktabs}%
\usepackage{algorithm}%
\usepackage{algorithmicx}%
\usepackage{algpseudocode}%
\usepackage{listings}%
\usepackage{url}
\usepackage[breaklinks=true, colorlinks=true, linkcolor=blue, citecolor=blue, urlcolor=blue]{hyperref} 
\usepackage{orcidlink}

\newcommand\blfootnote[1]
{%
  \begingroup
  \renewcommand\thefootnote{}\footnote{#1}%
  \addtocounter{footnote}{-1}%
  \endgroup
}

\begin{document}
\title{Automated Educational Question Generation at Different Bloom’s Skill Levels using Large Language Models: Strategies and Evaluation}




\titlerunning{AEQG at Different Bloom’s Skill Levels using LLMs}
%
\author{Nicy Scaria\inst{1}\orcidlink{0009-0004-8699-0312}, Suma Dharani Chenna\inst{1,2}\orcidlink{0009-0008-3260-3953}, Deepak Subramani\inst{1}\orcidlink{0000-0002-5972-8878}}
\authorrunning{N. Scaria et al.}
%
\institute{Computational and Data Sciences, Indian Institute of Science, India \and
School of Computer Science and Engineering, VIT-AP University, India \\
\email{\{nicyscaria,deepakns\}@iisc.ac.in, sumadharanichenna@gmail.com}}
%

\maketitle   

\vspace{-0.5cm}
\begin{abstract}
Developing questions that are pedagogically sound, relevant, and promote learning is a challenging and time-consuming task for educators. Modern-day large language models (LLMs) generate high-quality content across multiple domains, potentially helping educators to develop high-quality questions. Automated educational question generation (AEQG) is important in scaling online education catering to a diverse student population. Past attempts at AEQG have shown limited abilities to generate questions at higher cognitive levels. In this study, we examine the ability of five state-of-the-art LLMs of different sizes to generate diverse and high-quality questions of different cognitive levels, as defined by Bloom’s taxonomy. We use advanced prompting techniques with varying complexity for AEQG. We conducted expert and LLM-based evaluations to assess the linguistic and pedagogical relevance and quality of the questions. Our findings suggest that LLMs can generate relevant and high-quality educational questions of different cognitive levels when prompted with adequate information, although there is a significant variance in the performance of the five LLMs considered. We also show that automated evaluation is not on par with human evaluation. 
\end{abstract}

\blfootnote{This is a preprint. The Version of Record of this contribution is published in International Conference on Artificial Intelligence in Education (LNAI,volume 14830), and is available online at \url{https://doi.org/10.1007/978-3-031-64299-9_12}}

\vspace{-1.25cm}
\section{Introduction}

Transformer-based pre-trained large language models developed in recent years have drastically improved the quality of natural language generation (NLG) tasks \cite{zhang2022survey}. With an exponential increase in training data and model size, these models can generate complex text with human expert-level quality. The release of OpenAI’s ChatGPT made LLMs accessible to a wider audience who are not experts in natural language processing (NLP), allowing them to use them for their daily tasks. The language models are tuned to follow the user instructions through instruction-tuning \cite{zhang2022survey}. They have zero-shot capabilities \cite{kojima2022large}, which means that if you prompt the LLM with detailed task descriptions, the model will create meaningful outputs. These LLMs have the potential to be used in different ways in education \cite{kasneci2023chatgpt}, including the creation of personalized content, assessments, and feedback.

High-quality assessments enable learners to deeply engage with the subject and relate their learning to the real world. Assessments that focus on different cognitive skills as defined in Bloom’s taxonomy levels \cite{blooms} (described in Table~\ref{bloom}) help educators identify the gaps in student learning. This information allows them to adapt their teaching to better support students and also helps students understand their strengths and weaknesses. However, creating such assessments requires significant time and effort from educators \cite{q3}. Automated Educational Question Generation (AEQG) systems reduce the effort and cognitive load on teachers. Past research on AEQG methods required context information for the models to generate high-quality questions. Educational information is available from multiple sources today and choosing the right resource is challenging. 

\textbf{Related Work}: In the pre-LLM era, AQG research focused mainly on generating questions using question-answer datasets such as SQuAD 2.0, and NQ. These data sets contained a context and an answer for which the question had to be created \cite{zhang2021review}. However, the limited availability of public datasets impeded the progress of AQG systems capable of producing good quality questions. Recent research in question generation is focused on using pretrained or fine-tuned LLMs for the process. Encoder decoder models, such as the Text-To-Text Transfer Transformer (T5) and decoder-only models such as GPT3, along with context information, were used to generate questions \cite{nguyen2022towards}. Pre-training these models with educational text also improved the quality of the questions generated \cite{qg1}. 

Recent research has shown promising results in evaluating the quality of machine-generated content using LLMs using Chain-of-Thought (CoT) prompting on different evaluation criteria. G-EVAL \cite{Geval}, an evaluator model based on GPT4, significantly outperformed previous models and aligned with human judgments on summarization tasks. However, the results of some studies that used fine-tuned GPT3 models to evaluate the pedagogical quality of machine-generated questions were unsatisfactory \cite{moore2022assessing,qg1}. Human expert or crowd evaluations have been extensively used to analyze the pedagogical quality of machine-generated questions \cite{ped_quality,ped_quality1}.
 
The questions generated by most AQG models generally test lower-order skills \cite{ushio-etal-2022-generative} or create questions that have answers directly mentioned in the text \cite{zhang2021review}. These questions are not enough to test the higher-order cognitive skills of students. Bloom’s taxonomy \cite{blooms} serves as a guide for educators to generate questions to test different cognitive skills. Recent work \cite{sridhar2023harnessing} has used GPT4 to develop course material based on Bloom’s taxonomy. 


\begin{table}[h]
\caption{Revised Bloom’s taxonomy \cite{blooms} in ascending order in the cognitive dimension}
\label{bloom}
\centering
\begin{tabular}{p{2.15cm}p{9.85cm}}
\toprule                 
Bloom’s level     & Description \\
\midrule
Remember & Retrieve relevant knowledge from long-term memory. \\
Understand & Construct meaning from instructional messages, including oral, written, and graphic communication.\\
Apply  & Carry out or use a procedure in a given situation. \\
Analyze  & Break material into foundational parts and determine how parts relate to one another and the overall structure or purpose \\
Evaluate  & Make judgments based on criteria and standards. \\ 
Create & Put elements together to form a coherent whole; reorganize into a new pattern or structure. \\
\bottomrule
\end{tabular}
\end{table}


\subsection{Objective and Research Questions}
\label{sec:RQ}
Our approach utilizes the knowledge of the content inherently present in LLMs along with the addition of technical information on the question generation process in the prompt to generate educational questions.
Although LLMs excel in various downstream tasks, they produce errors and inconsistencies \cite{ji2023survey}, which can compromise the quality of the questions generated. This also varies significantly between different LLMs. Therefore, evaluating the quality of the questions generated by LLMs is essential. While metrics such as the BLEU score or perplexity can assess machine-generated questions, they typically only examine linguistic characteristics \cite{wang2022towards}. 
In the present work, we perform a manual expert evaluation using the services of two educators in the AEQG topic’s domain and an automated LLM evaluation using an LLM that is not employed for AEQG.

We used zero-shot and few-shot techniques and CoT prompting to generate questions for a graduate-level data science course using LLMs of different sizes. Five different prompt strategies of varying complexity were used to create these questions. Then, we performed a manual expert evaluation using the services of two educators in the AEQG topic’s domain and an automated LLM evaluation using an LLM that is not employed for AEQG. The evaluation was performed on a nine-item rubric to assess their linguistic and pedagogical quality \cite{horbach} by experts and the LLM. The LLM evaluation is performed as a zero-shot classification task through a specially designed prompt.

Specifically, we investigated answers to the following research questions.

\noindent \textbf{RQ1}: Can instruction fine-tuned modern LLMs create high-quality and diverse educational questions at different cognitive levels based on Bloom’s taxonomy? 

\noindent \textbf{RQ2}: Does the size of the LLM significantly impact the model’s performance in educational question generation? 

\noindent \textbf{RQ3}: How does the amount of information provided in the prompt affect the quality of the questions generated? 

\noindent \textbf{RQ4}: Can LLMs create questions that are relatable to a specific population or context? 

\noindent \textbf{RQ5}: Can instruction fine-tuned LLMs evaluate generated educational questions effectively, similar to human evaluators, when given the same instructions? 


In what follows, we first discuss the methodology. Then, we present and analyze the results before concluding with directions for future research. 
\section{Methodology}


Our study consists of two parts. We use modern LLMs for AEQG in the first part through various prompting strategies. In the second part, we perform human evaluation and LLM evaluation.

\subsection{Language Models}
\label{lang_models}
Training a language model from scratch or fine-tuning the available models is expensive due to constraints in the availability of educational data and the cost associated with training. Therefore, we used a mix of open-source and proprietary state-of-the-art LLMs for the study. The models used for question generation are Mistral (Mistral-7B-Instruct-v0.1), Llama2 (Llama-2-70b-chat-hf), Palm 2 (chat-bison-001), GPT-3.5 (gpt-3.5-turbo-0613), and GPT-4 (gpt-4-0613). Among these, Mistral has 7 billion parameters, and GPT models are rumored to have trillions of parameters\footnotemark[1]. For LLM-based evaluation of the questions, we used Gemini Pro (gemini-pro). 


\footnotetext[1]{The information about the number of parameters of the model is not publicly available}

\subsection{Question Generation}
\label{sec:promptstrategy}
We used the five LLMs mentioned in Sect.~\ref{lang_models} to generate questions of different cognitive levels. A higher temperature setting in LLMs results in a varied and unpredictable text, while a lower temperature setting makes the model output more deterministic and repetitive. Thus, we set the temperature of the LLMs at 0.9 to promote variety and diversity in the generated questions. 
\subsubsection{Content}: The educational questions were generated for a graduate-level data science course comprising topics ranging from traditional machine learning algorithms, such as linear regression, to advanced topics in natural language processing, such as prompt engineering. We did not provide domain-specific information or context on these topics to the models for question generation. This approach was guided by the hypothesis that these models, trained using large amounts of recent Internet data, would possess inherent knowledge related to these contemporary course topics. 

\subsubsection{Prompt Design}: In the present study, we generated questions by instructing the models with five prompt styles/strategies (PS1 to PS5), each differing in complexity. These prompts followed specific techniques of pattern reframing, itemizing reframing, and assertions to make it easier for the instruction fine-tuned LLMs to follow the specific instructions \cite{reframing}. Furthermore, the prompts encouraged the model to incorporate Indian-specific examples or context within the question to ensure relevance for Indian students. The first set of prompts (PS1) consisted solely of these core instructions. In contrast, subsequent sets progressively added more specific information and instructions in the prompt to further refine the quality and relevance of the generated questions. 

There were mainly three significant additions to the prompts. First, the prompts were augmented with CoT instructions to make the LLM think sequentially about how to proceed with the task. CoT prompting has been shown to improve the quality of LLM-generated content \cite{cot}. In the prompt, the LLM was also given the persona of a graduate-level university course instructor creating questions for their students \cite{reframing}. Second, the questions included definitions of the six cognitive levels of the revised Bloom’s taxonomy. This approach was taken under the hypothesis that augmenting the LLM with explicit knowledge of the revised Bloom's taxonomy levels would enhance the quality of the generated questions. Third, an expert-crafted example was provided for each Bloom’s taxonomy level. This few-shot approach, proven effective in different generative models \cite{prompts}, leverages human-crafted questions to guide the LLM’s understanding of how the questions need to be framed and, in turn, enhance its question generation capabilities. Using these three, we created four different prompts: (PS2) CoT prompt with skill explanation, (PS3) CoT prompt with example questions, (PS4) CoT prompt with skill and example questions, and (PS5) CoT prompt with skill, skill explanation, and example questions.

We gave the same prompt to all LLMs in a specific combination, except for the topic-specific variables corresponding to each course topic, as given in the repository\footnotemark[2]. Each LLM generated six questions, one for each level of Bloom's taxonomy corresponding to the 17 course topics. Each model generated 102 questions, resulting in 510 questions for one combination, making it 2550 for the five prompt combinations. 
\footnotetext[2]{\url{https://github.com/nicyscaria/AEQG_Blooms_Evaluation_LLMs}}

\subsection{Human Evaluation}
\label{human-eval}

Two experts evaluated the AEQG questions. Both experts deeply understand data science concepts and have experience teaching the subject to large graduate classes. The experts assessed the relevance and quality of the questions based on a nine-item rubric (Table~\ref{rubric}; a modified version of \cite{horbach}). 

The experts were presented with LLM-generated questions in random order without information other than the course topic on the prompt corresponding to the question. The experts hierarchically assessed each question, starting from the top of the rubric to the bottom. In the evaluation, each group of the evaluation criteria, as indicated in Table~\ref{rubric}, has a stopping point. This structured approach streamlines the evaluation process, minimizing the overall time and effort required for expert annotation. If \textit{Understandable} is marked `no’, then none of the subsequent items are evaluated for that question and are automatically marked as `NA’, indicating not applicable. This design choice reflects the underlying principle that further evaluation of a question that is not understandable does not make sense. Similarly, within group 2, if the answer to \textit{clear} is `no’, then the evaluation stops, and the remaining items are automatically marked as `NA’. Additionally, if the response to \textit{Clear} is `more\_or\_less’, then the \textit{Rephrase} criteria of group 3 should be marked, and the evaluator should rephrase the question for clarity. The rephrased version of the question will be used for further evaluation. If the answer to \textit{Answerable} in group 3 is `no', the evaluation is stopped, and the remaining items of the rubric are marked `NA’. In group 4, if either \textit{Central} or \textit{WouldYouUseIt} is marked `no’, then the evaluation is stopped, and the \textit{Bloom’sLevel} criteria is labeled `NA’; otherwise, the experts select the Bloom’s level for the question and concludes the evaluation of one question. The process is repeated for all the questions.

\begin{table}[H]
\caption{Hierarchical nine-item rubric used to evaluate questions generated by LLMs}
\label{rubric}
\centering
\begin{tabular}{p{2.5cm}p{9.5cm}}
\toprule                 
Rubric item & Definition and response option\\
\midrule
Understandable & Could you understand what the question is asking? \textit{(yes/no)}\\
\midrule
TopicRelated  & Is the question related to the topic given in the prompt? \textit{(yes/no)}\\
Grammatical & Is the question grammatically well-formed? \textit{(yes/no)}\\
Clear  & Is it clear what the question asks for? \textit{(yes/more\_or\_less/no)}\\
\midrule
Rephrase  & Could you rephrase the question to make it clearer and error-free? \textit{(yes/no)}\\
Answerable  & Can students answer the question with the information or context provided within? \textit{(yes/no)}\\
\midrule
Central & Do you think being able to answer the question is important to work on the topic given in the prompt? \textit{(yes/no)}\\
WouldYouUseIt  & If you were a teacher teaching the course topic would you use this question or the rephrased version in the course? \textit{(yes/maybe/no)}\\
\midrule
Bloom’sLevel & What is the Bloom’s skill associated with the question? \textit{(remember, understand, apply, analyze, evaluate, and create)} \\
\bottomrule
\end{tabular}
\end{table}

The AEQG questions were considered high quality if they met the following criteria: (1) experts marked `yes’ for \textit{Understandable}, \textit{Grammatical}, \textit{Clear}, and \textit{Answerable}; (2) received a `yes’ or `maybe’ for \textit{WouldYouUseIt}; or (3) being marked `yes’ for `more\_or\_less’ in the \textit{Clear} criteria and subsequently marked `yes’ for \textit{Rephrase}. Furthermore, we utilized \textit{Bloom’sSkill} to understand whether the LLM adheres to the instructions provided in the prompt. The LLM adhered to the instructions provided if the \textit{Bloom’sSkill} labels of the experts match the Bloom’s skill level on the prompt.

Experts perceive the questions generated by LLMs in different ways. This perception is influenced by various factors, such as their preference for writing, personal assumptions, prior knowledge, and attention to detail \cite{amidei2018rethinking}. Therefore, it is essential to have a measure to ensure the consistency of the expert evaluation. We measure inter-rater reliability using percentage agreement and Cohen’s Kappa $\kappa$ \cite{mchugh2012interrater}. For the ordinal metrics, \textit{Clear}, \textit{WillYouUseIt} and \textit{Bloom’sLevel}, we use quadratic weighted Cohen’s $\kappa$ \cite{cohen1968weighted} instead of simple Cohen’s $\kappa$ to penalize situations with a significant rating difference. 

Along with the metrics discussed above, we explored the ability of LLMs to create questions relatable to a specific population or context, which, in this case, is India. To understand this, we curated and analyzed the contexts and themes specific to India that came up in the questions.

\subsection{Automated Evaluation} \label{auto_eval}
Clearly, the above evaluation by a human expert is a laborious process. Automated evaluation offers a scalable and efficient alternative for assessing large-scale educational content. We use Gemini Pro for the LLM-based evaluation of the generated questions. To ensure deterministic behavior, we set the decoding temperature of the model to 0. In automated evaluation, we assess the quality of LLM-generated questions without reference questions using the same criteria outlined in Sect.\ref{human-eval}. Recent studies have demonstrated the ability of LLMs to perform a reference-free evaluation for a variety of NLG tasks \cite{Geval,chatgpt_evaluate}. The prompt used in the prompt-based evaluator consisted of two components: (1) a detailed description of the evaluation criteria, evaluation instructions (instructions to evaluate the questions in a hierarchical manner as discussed in Sect.\ref{human-eval}) along with the question and the course topic for which the question was generated, and (2) CoT instructions describing the evaluation steps, along with providing the evaluator LLM the persona of a graduate-level data science course instructor. The detailed prompt template can be found on github.

In addition to automated evaluation, we assessed the linguistic quality of the AEQG questions using a diversity measure based on the Paraphrase In N-gram Changes (PINC) score \cite{pinc}. 
\begin{equation}
\label{PINC_formula}
    PINC(s, c) = \frac{1}{N} \sum_{n=1}^{N} \left( 1 - \frac{|n\text{-}gram_s \cap n\text{-}gram_c|}{|n\text{-}gram_c|} \right)
\end{equation}
PINC score is generally used to measure the novelty of n-grams in the automatically generated paraphrase of a sentence. In our case, we wanted to check if the questions generated by an LLM for a specific Bloom’s skill use the same structure or words on different topics. For that, we considered every question generated for a specific skill by the model as the source and calculated the PINC score considering every other question of the same skill and by the same model as the candidate question. Finally, the average of these scores was calculated for the AEQG questions generated by each LLM. A higher average PINC score indicates considerable diversity in the AEQG task.

\section{Results and Analysis}
\label{sec:results}
We will be releasing the dataset, `DataScienceQ'\footnotemark[4] containing 2550 questions generated for the present study. First, we present and analyze the results of the expert evaluation of these 2550 AEQG questions.
\footnotetext[4]{\url{https://github.com/nicyscaria/AEQG_Blooms_Evaluation_LLMs}}
We start by examining the agreement between experts on their evaluation using the percentage agreement and modified Cohen’s $\kappa$ values (Table~\ref{expert_eval}). The percentage agreements and Cohen’s $\kappa$ values are calculated only for questions not labeled `NA’ as discussed in Sect.\ref{human-eval}. The values in the table indicate that there is substantial agreement between experts on different evaluation criteria. The agreement is highest for the criteria \textit{TopicRelated}, \textit{Grammatical}, and \textit{Central}. As expected, subjective criteria like \textit{Rephrase} and \textit{WouldYouUseIt} have the lowest agreement. In our analysis, an AEQG question is considered as ``High Quality'' only when both evaluators rate it as ``High Quality''. To assess alignment with Bloom's taxonomy, the subsequent analysis focused only on these ``High Quality'' questions. In what follows, we discuss the results for each research question stated in Sect.~\ref{sec:RQ}. Table~\ref{tab:performance_llms} presents the key evaluation metrics (Sect.~\ref{human-eval}) for the analysis. Quality is measured as the percentage of AEQG questions that were selected as ``High Quality'' and skill is measured as the percentage of ``High Quality'' questions judged by experts to be at the same Bloom's taxonomy levels as the one given in the prompt to the LLM for the AEQG task.

\begin{table}[H]
\caption{Expert inter-annotator agreement on the nine-item hierarchical rubric for AEQG questions.}
\label{expert_eval}
\customsize
\begin{tabular*}{\textwidth}{@{\extracolsep\fill}lcccccc}
\toprule
& \multicolumn{2}{c}{Simple prompt} & \multicolumn{2}{c}{CoT \& skill explanation} & \multicolumn{2}{c}{CoT \& example}\\
\cmidrule{2-3}\cmidrule{4-5} \cmidrule{6-7}
Rubric item & \% agree & $\kappa$ & \% agree & $\kappa$ & \% agree & $\kappa$\\
\midrule
Understandable  & 99.60\% & 0.67 & 99.61\% & 0.80 & 100.00\% & 1.00\\
TopicRelated  & 99.80\% & 0.95 & 99.80\% & 0.80 & 98.80\% & 0.93\\
Grammatical   & 100.00\% & 1.00 & 100.00\% & 1.00 & 100.00\% & 1.00\\
Clear  & 91.75\% & 0.67 & 97.42\% & 0.62 & 94.30\% & 0.61\\
Rephrase  & 90.38\% & 0.76 & 90.91\% & 0.62 & 80.77\% & 0.59\\
Answerable  & 93.41\% & 0.65 & 96.47\% & 0.69  & 94.75\% & 0.61\\
Central  & 97.42\% & 0.78 & 99.77\% & 0.98 & 99.77\% & 0.94\\
WouldYouUseIt  & 89.92\% & 0.58 & 94.82\% & 0.66 & 96.81\% & 0.69\\
Bloom’sLevel  & 82.07\% & 0.83 & 88.14\% & 0.90 & 90.00\% & 0.89\\
\midrule
\end{tabular*}
\begin{tabular*}{\textwidth}{@{\extracolsep\fill}lcccc}
& \multicolumn{2}{c}{CoT, skill, and example} & \multicolumn{2}{c}{CoT, skill, skill explanation and example} \\ 
\cmidrule{2-3} \cmidrule{4-5}
Rubric item & \% agree & $\kappa$ & \% agree & $\kappa$ \\ 
  \midrule
Understandable  & 100.00\% & 1.00 & 100.00\%     & 1.00     \\
TopicRelated    & 99.80\% & 0.99 & 99.20\%      & 0.96     \\
Grammatical     & 100.00\% & 1.00 & 100.00\%     & 1.00     \\
Clear           & 93.30\% & 0.53 & 92.74\%      & 0.66     \\
Rephrase        & 92.31\% & 0.73 & 75.00\%      & 0.53     \\
Answerable      & 96.54\% & 0.81 & 93.89\%      & 0.68     \\
Central         & 100.00\% & 1.00 & 100.00\%     & 1.00     \\
WouldYouUseIt   & 95.64\% & 0.85 & 95.29\%      & 0.82     \\
Bloom’s Level   & 90.36\% & 0.93 & 92.65\%      & 0.93     \\ 
\bottomrule
\end{tabular*}
\end{table}

\noindent \textbf{RQ1: Can instruction fine-tuned modern LLMs create high-quality and diverse educational questions at different cognitive levels based on Bloom’s taxonomy?} Among all AEQG questions from the different LLMs and prompting strategies, 78\% were rated as ``High Quality'' and among these 65.56\% were rated to match the intended skill level by both human raters (Table~\ref{tab:performance_llms} `Overall' Columns and `Overall' row). The temperature of all LLMs was set at 0.9 to promote textual diversity, resulting in a PINC score average of 0.92. These findings suggest that instruction fine-tuned LLMs demonstrate considerable potential to generate diverse and high-quality educational questions at different cognitive levels based on Bloom’s taxonomy. For PS1-PS5, 72.55\%, 70.58\%, 71.56\%, 86.27\%, and 89.02\% of the questions were identified as ``High Quality" for Mistral 7B, Llama2 70B, Palm 2, GPT 3.5 and GPT 4 respectively. Similarly, 60\%, 61.67\%, 60\%, 74.41\%, and 70.04\% of the questions followed adhered to Bloom's taxonomy level given by experts respectively.

\vspace{0.15cm}

\noindent \textbf{RQ2: Does the size of the LLM significantly impact the model’s performance in educational question generation?} Table~\ref{tab:performance_llms} presents the performance metrics of the five LLMs for the five sets of prompts. For quality and adherence to Bloom’s taxonomy levels, GPT 4 and GPT 3.5 emerged as the top performers. 
Palm 2, despite its larger size compared to Mistral 7B and LLama2 70B, demonstrated wide variance in the quality of the AEQG task for different prompt strategies. Palm 2 has only 36.99\% Skill matching in a detailed and complex prompt (PS5), but it scores 70.51\% in the PS3 prompt strategy. For PS5 prompts, the Mistral 7B model performs better than the Llama2 70B model, which is counterintuitive. The reason for this performance difference could be due to the way these models process a long prompt. Thus, no clear pattern exists between the model size and AEQG performance.
\begin{table}[H]
\caption{Performance of the LLMs in the AEQG task. For each model and set of prompts PS1-PS5, the percentage of questions that are of high quality (Quality), adherence to Bloom’s taxonomy level (Skill), and the PINC score are presented.}
\label{tab:performance_llms}
\customsize
\begin{tabular*}{\textwidth}{@{\extracolsep\fill}lccccccccc}
\toprule
& \multicolumn{3}{c}{PS1: Simple prompt} & \multicolumn{3}{c}{PS2: CoT \& skill explanation} & \multicolumn{3}{c}{PS3: CoT \& example}\\
\cmidrule{2-4}\cmidrule{5-7} \cmidrule{8-10}
LLM & Quality & Skill & PINC & Quality & Skill & PINC & Quality & Skill & PINC\\
\midrule
Mistral 7B  & 70.59\%  & 56.94\%    & 0.94 &  75.49\% & 68.83\%  & 0.95 & 78.43\% & 45.00\% & 0.94\\
Llama 2 70B & 73.53\%  & 57.33\%    & 0.94 &  75.49\% & 71.43\% & 0.93 & 77.45\% & 58.23\% & 0.93\\
Palm 2   & 61.76\%     & 55.56\%    & 0.93 &  73.53\% & 65.33\% & 0.94 & 76.47\% & \textbf{70.51\%} & 0.93\\
GPT 3.5  & 69.61\%     & \textbf{81.69\%}   & 0.94 &  \textbf{94.12\%} & \textbf{89.58\%} & 0.89 & 89.22\% & 64.84\% & 0.92\\
GPT 4  & \textbf{75.49\%}       & 74.03\%    & 0.93 &  87.25\% & 82.02\% & 0.92 & \textbf{91.18\%} & 65.59\% & 0.92\\
Overall  & 70.20\%       & 66.36\%    & 0.93 &  81.18\% & 76.32\% & 0.93 & 82.55\% & 61.04\% & 0.93\\
\midrule
\end{tabular*}
\begin{tabular*}{\textwidth}{@{\extracolsep\fill}lccccccccc}
& \multicolumn{3}{c}{PS4: CoT, skill, \&} & \multicolumn{3}{c}{PS5: CoT, skill, skill} & \multicolumn{3}{c}{Overall}\\
& \multicolumn{3}{c}{ example} & \multicolumn{3}{c}{explanation \& example} & \multicolumn{3}{c}{}\\
\cmidrule{2-4}\cmidrule{5-7} \cmidrule{8-10} 
LLM & Quality & Skill & PINC & Quality & Skill & PINC & Quality & Skill & PINC\\
  \midrule
  Mistral 7B & 71.57\% & 71.23\% & 0.93 & 66.67\% & 58.82\% & 0.94 & 72.55\% & 60.00\%  & 0.94\\
Llama 2 70B & 77.45\% & 73.42\% & 0.91 & 49.02\% & 40.00\% & 0.93 & 70.58\% & 61.67\% & 0.93\\
Palm 2 & 74.51\% & 69.74\% & 0.93 & 71.57\% & 36.99\% & 0.93 & 71.56\% & 60.00\% & 0.93\\
GPT 3.5 & 87.25\% & 73.03\% & 0.90 & \textbf{91.18\%} & \textbf{59.14\%} & 0.91 & 86.27\% & \textbf{73.41\%} & 0.91\\
GPT 4 & \textbf{96.08\%} & \textbf{75.51\%} & 0.92 & 95.10\% & 56.64\% & 0.90 & \textbf{89.02\%} & 70.04\% & 0.92\\
Overall & 81.37\% & 72.77\% & 0.92 & 74.71\% & 51.18\% & 0.92 & 78.00\% & 65.56\% & 0.93\\
\bottomrule
\end{tabular*}
\end{table}

\begin{figure}[H]%
\centering
\includegraphics[width=\textwidth]{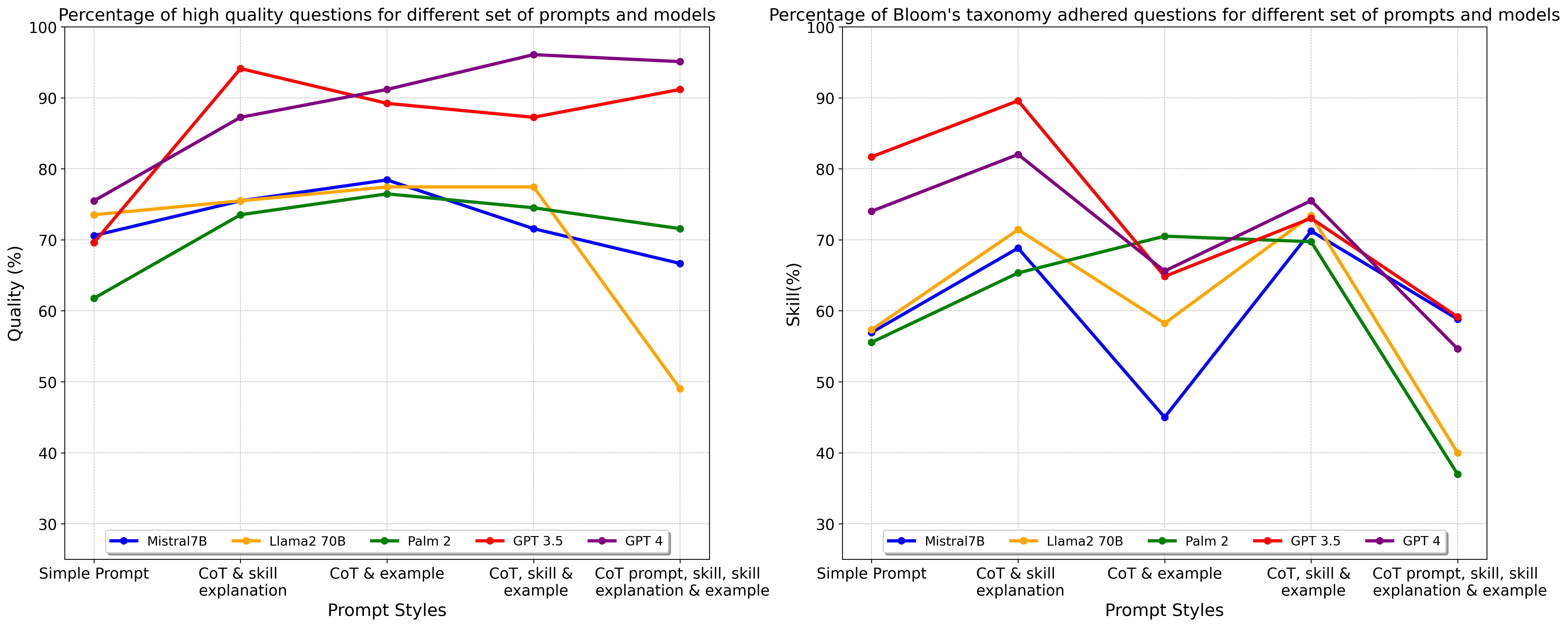}
\caption{Quality and Skill of prompting strategies for AEQG by different LLMs.}\label{fig:question_quality}
\end{figure}

\noindent \textbf{RQ3: How does the amount of information provided in the prompt affect the quality of the questions generated?} It is observed that the simple prompt (PS1) performed poorly in the quality of the questions generated (Table~\ref{tab:performance_llms}). Overall, the performance improved with the addition of more information to the prompt. However, the amount of improvement varied between the five LLMs. Figure~\ref{fig:question_quality} shows that for Mistral, Llama 2 and Palm 2, PS3 gave the highest quality questions, with PS4 being close behind. PS4 gave the highest skill match for these three models, while the skill match for PS3 was low. Interestingly, PS5, which is the most complicated prompt used, reduced both quality and skill for these three models, indicating that while information enrichment improves AEQG, too much information in the prompt can be counterproductive. The GPT models also gave good performance for PS4, but their performance for PS2 to PS5 are more or less the same with respect to quality of questions. These two models did well in terms of skill match for PS2-PS4, and similar to the other three LLMs, the PS5 skill scores drop significantly. Our results indicate that a CoT prompt with a description of the skill and an example question performs best for AEQG.


\begin{figure}[H]%
\centering
\includegraphics[width=0.75\textwidth]{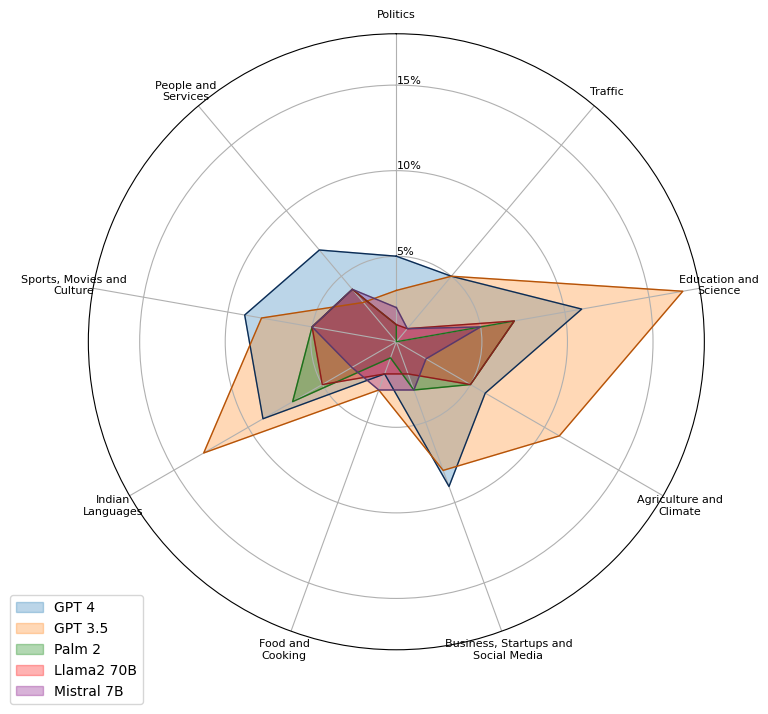}
\caption{Frequently repeated Indian contexts in the AEQG  questions.}\label{themes_fig}
\end{figure}

\noindent \textbf{RQ4: Can LLMs create questions that are relatable to a specific population or context?} In our study, LLMs were asked to create questions that are relatable to students in India. Nine recurring themes that are specific to India emerged (Fig.~\ref{themes_fig}). These included questions on Bollywood movies, traffic challenges in Indian cities and their mitigation using computer vision techniques, and the Indian educational system. Given India's significant dependence on agriculture, many questions focused on climate, crop yield, crop diseases, and cropping patterns. In addition, numerous questions, specifically on the topic of natural language processing, used Indian languages as examples. Interestingly, the Indian language text generated by open-source models, Mistral 7B and Llama2 70B, was often inaccurate and poor quality. From Fig.~\ref{themes_fig}, it is also clear that compared to GPT 4 and GPT 3.5, the recurring themes occur less in the questions generated by other models. Some questions exhibited a tendency to unnecessarily specify India when the context was general. The expert evaluators rephrased these questions. Furthermore, some questions reflect certain cultural generalizations about India that are not necessarily true, as evaluators have indicated.

\noindent \textbf{RQ5: Can instruction fine-tuned LLMs evaluate generated educational questions effectively, similar to human evaluators, when given the same instructions?} We conducted an LLM-based evaluation to analyze the quality and adherence of machine-generated questions to different cognitive levels on the nine-item rubric in addition to the expert evaluation. We used Gemini Pro (gemini-pro), an LLM that is different from the five used for the AEQG task, for the evaluation (detailed methodology in Sect.~\ref{auto_eval}). The results of the evaluation are given in Table~\ref{performance_llms_gemini}. There is a significant discrepancy between LLM-based and expert evaluations. Interesting discrepancies emerged between Gemini Pro and expert evaluations, with Palm 2 excelling in automated evaluation, but underperforming in expert evaluation. In the Gemini Pro evaluation, even Llama 2 70B and Mistral 7B also performed better in some cases. Our automated evaluation using Gemini Pro revealed a tendency of the model to classify most machine-generated questions as belonging to the `Apply' or `Analyze' levels on \textit{Bloom'sLevel}. The observed performance dip of the LLMs in adhering to the evaluator-LLM's Bloom's level can be attributed to this fact. This result indicates that extreme caution must be exercised when using LLMs for automated evaluation of generative tasks. In the future, our data set can be used to improve the automated evaluation of the questions as well. 

\begin{table}[H]
\caption{Automated evaluation of AEQG questions: percentage of high-quality questions and adherence to Bloom’s taxonomy level given by Gemini Pro.}
\label{performance_llms_gemini}
\customsize
\begin{tabular*}{\textwidth}{@{\extracolsep\fill}lcccccc}
\toprule
& \multicolumn{2}{c}{Simple prompt} & \multicolumn{2}{c}{CoT \& skill explanation} & \multicolumn{2}{c}{CoT \& example}\\
\cmidrule{2-3}\cmidrule{4-5} \cmidrule{6-7}
LLM & Quality & Skill & Quality & Skill & Quality & Skill\\
\midrule
Mistral 7B  & 82.35\%  & \textbf{53.98\%}    &   61.76\% & 44.44\%  &  70.59\% & 19.44\%  \\
Llama 2 70B & 79.41\%  & 40.02\%    &   65.69\% & 43.28\% &  \textbf{80.39\%} & 40.24\% \\
Palm 2   & 69.61\%     & 39.43\%    &  65.69\% & \textbf{49.25\%} &  78.43\% & \textbf{40.00\%} \\
GPT 3.5  & \textbf{82.35\%}     & 48.15\%    &   67.65\% & 34.78\% &  75.49\% & 28.24\% \\
GPT 4  & 80.39\%       & 38.09\%    &   \textbf{75.49\%} & 35.06\% &  77.45\% & 37.97\%  \\
\midrule
\end{tabular*}
\begin{tabular*}{\textwidth}{@{\extracolsep\fill}lcccc}
\toprule
& \multicolumn{2}{c}{CoT, skill, and example} & \multicolumn{2}{c}{CoT, skill, skill explanation and example.}\\
\cmidrule{2-3}\cmidrule{4-5}
LLM & Quality & Skill  & Quality & Skill \\
  \midrule
  Mistral 7B & 58.82\% & 40.00\% & 55.88\% & 31.57\% \\
Llama 2 70B & 62.75\% & 39.06\% & 53.92\% & 34.54\% \\
Palm 2 & \textbf{77.45\%} & \textbf{44.30\%} & \textbf{69.61\%} & 36.62\% \\
GPT 3.5 & 66.67\% & 33.82\% & 51.96\% & 30.18\% \\
GPT 4 & 73.53\% & 40.00\% & 66.67\% & \textbf{38.24\%} \\
\bottomrule
\end{tabular*}
\end{table}

\section{Discussion and Conclusion}


Our study demonstrates that LLMs can produce high-quality and diverse educational questions aligned with Bloom's taxonomy, requiring minimal input from educators, but the performance varies based on the size of the model and the prompt used to generate these questions. Larger proprietary models like GPT 4 and GPT 3.5 outperform smaller open-source models across all the metrics in expert evaluation, but the same does not hold for the Palm 2 model. While adding a lot of information (skill explanation, example questions, and CoT instructions) significantly reduced the performance of the LLMs, particularly for open-source models, optimal results were achieved with prompts including CoT instructions paired with either skill, skill explanation, or example questions. CoT instructions, with examples, resulted in more high-quality questions while compromising on the adherence to Bloom's skill. On the other hand, Bloom's skill explanations with CoT instructions slightly reduced the number of high-quality questions but significantly boosted the performance of adherence to Bloom's skill. The questions generated often incorporated contextually relevant Indian contexts, although some instances exhibited generalizations about India. 

In our research, the evaluation of 2550 questions took a considerable amount of time and effort from expert evaluators. Although attention was paid to making the evaluation process objective, experts' decisions can still be subjective depending on who is evaluating them. However, the LLM-based evaluation proved to be less effective in our case. There was a considerable difference across all metrics in the case of the Gemini Pro evaluation compared to the expert evaluation. Interestingly, in the Gemini Pro evaluation, Palm 2 outperformed other models on different evaluation metrics, even though expert evaluation suggested otherwise. This discrepancy might stem from the fact that Palm 2 and Gemini Pro are both Google's models, potentially sharing similar training data or methodologies. We found that our evaluator model is suboptimal and does not align with the expert evaluation. This could be due to the lack of such examples that these models would have seen during their training. This requires the training of the LLM on evaluation datasets on specific subjects and evaluation metrics to make it robust for evaluation. This is a potential future direction of research. 

Our approach used the inherent knowledge of the content possessed by the LLMs on the topic for AEQG. This revealed limitations in their understanding of specific domains. For example, Mistral 7B and Llama2 70B struggled on the topics of "prompt engineering", producing questions related to general engineering. Another interesting observation in the study was the inability of most models to generate high-quality questions at the `Create' level of Bloom's taxonomy. Thus, the present paper can be extended to use existing resources from the Internet or course material through databases to improve the performance of the model in question generation. We did not study the generation of questions for topics other than data science content, and such an exploration using the methodology showcased here could be a potential future direction for research.

%
%
%
\bibliographystyle{splncs04}
\bibliography{references}
%




\end{document}